\documentclass[10pt,twocolumn,letterpaper]{article}
\pdfoutput=1

\usepackage{3dv}
\usepackage{times}
\usepackage{epsfig}
\usepackage{graphicx}
\usepackage{amsmath}
\usepackage{amssymb}
\usepackage{booktabs} 
\usepackage{subcaption}
\usepackage{float}
\usepackage[table,xcdraw]{xcolor}

\usepackage{pdfrender}
\usepackage{bm}

\usepackage[pagebackref=true,breaklinks=true,letterpaper=true,colorlinks,bookmarks=false]{hyperref}

\renewcommand{\vec}[1]{\boldsymbol{#1}}
\newcommand{\mat}[1]{\mathbf{#1}}

\newcommand{\tb}[0]{\textbf}

\newcommand{\normal}[0]{{\mathbf{n}}}

\newcommand{\ankl}[0]{\mathbf{x}_{l}}
\newcommand{\ankr}[0]{\mathbf{x}_{r}}
\newcommand{\lplane}[0]{L_{p}}

\newcommand{\pose}[0]{\vec{\theta}}
\newcommand{\shape}[0]{\vec{\beta}}

\newcommand{\joints}[0]{\mat{J}}
\newcommand{\jointsTwoD}[0]{\tilde{\mat{J}}}

\newcommand{\rot}[0]{\vec{R}}

\newcommand{\scale}[0]{\mat{s}}
\newcommand{\trans}[0]{\vec{t}}

\newcommand{\lossRep}[0]{L}

\threedvfinalcopy 

\def\threedvPaperID{0237} 

\ifthreedvfinal\fi
\begin{document}

\title{Body Size and Depth Disambiguation in \\Multi-Person Reconstruction from Single Images}

\author{
Nicolas Ugrinovic\hspace{0.4cm}Adria Ruiz\hspace{0.4cm}Antonio Agudo\hspace{0.4cm}\hspace{0.4cm}Alberto Sanfeliu\hspace{0.4cm}Francesc Moreno-Noguer 
\and
Institut de Robòtica i Informàtica Industrial, CSIC-UPC, Barcelona, Spain\\
{\tt\small \{nugrinovic,aruiz,aagudo,asanfeliu,fmoreno\}@iri.upc.edu
}}


\twocolumn[{%
\renewcommand\twocolumn[1][]{#1} %
\maketitle
\thispagestyle{empty}
\begin{center}
    \centering
    \vspace{-2mm}
    \includegraphics[width=.95\linewidth, trim={0cm 0.3cm 0cm 0}, clip = true]{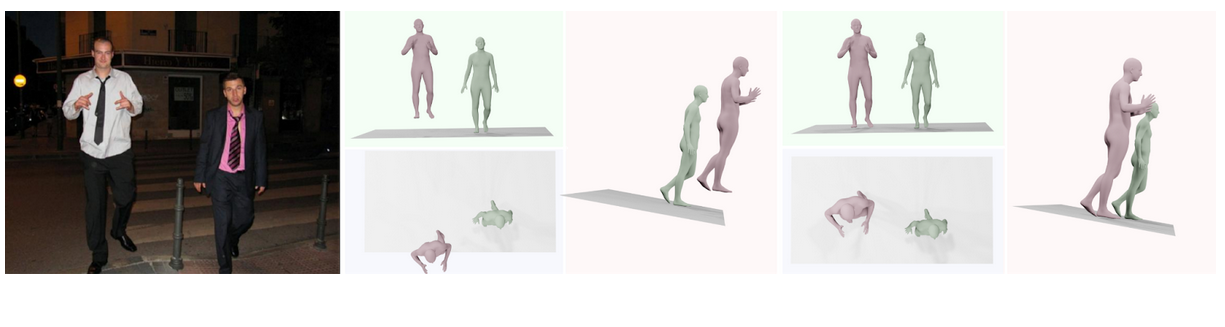}
        \put(-425,6){{{Input image}}}
        \put(-270,6){{{CRMH~\cite{jiang2020_multiperson}}}}
        \put(-95,6){{{Ours}}}
    \vspace{-0.9mm}
    \captionof{figure}{\small{We present a multi-person reconstruction approach that solves the inherent body size and depth ambiguity problem.  The figure shows the results of a state-of-the-art approach~\cite{jiang2020_multiperson} (center), and the results of our approach (right). Note that while the frontal view of both approaces is very similar, the side view in\cite{jiang2020_multiperson} depicts a large error. Our main contribution consist in proposing an optimization scheme that enforces the people to have their feet on the ground, which allows  disambiguating  body size from depth and producing accurate spatial arrangements. 
    }}\label{fig:teaser}
\end{center}
}]


\begin{abstract}\label{sec:abstract}

We address the problem of multi-person 3D body pose and shape estimation from a single image. While this problem  can be addressed by  applying  single-person approaches multiple times for the same scene, recent works have shown the advantages of building upon deep architectures that simultaneously reason about all  people in the scene in a holistic manner by enforcing, e.g., depth order constraints or minimizing  interpenetration among reconstructed bodies. However, existing approaches are still unable to capture the size variability of people caused by the inherent body scale and depth ambiguity. In this work, we tackle this challenge by devising a novel optimization scheme that learns the appropriate body scale and relative camera pose, by enforcing the feet of all people to remain on the ground floor. A thorough evaluation on MuPoTS-3D and 3DPW datasets demonstrates that our approach is able to robustly estimate the body translation and shape of multiple people while retrieving their spatial arrangement, consistently improving current state-of-the-art, especially in scenes with people of very different heights. Code can be found at: \small{\url{https://github.com/nicolasugrinovic/size_depth_disambiguation}}

\end{abstract}

\section{Introduction}\label{sec:intro}

While recent works on single-person 3D reconstruction have shown impressive results~\cite{bogo2016keep, guan2009estimating, kanazawa_hmr, kolotouros2019spin, kolotouros2019cmr, lassner2017unite, omran2018neural, smplx_2019, rong2021frankmocap, alldieck_learning_2019, tex2shape, saito2020pifuhd, onizuka2020tetratsdf,Corona_cvpr2021,Xu_pami2021}, the problem of simultaneously reconstructing multiple humans is still in its infancy. The straight-forward solution consists in regarding different people as independent instances and estimating the body shapes and poses one by one using a single-person approach. This strategy, however, may result in inconsistent spatial arrangements and erroneous poses of the reconstructed people. A few works have improved the coherence of the reconstructions by simultaneously reasoning about all people in the image, either for estimating their body poses~\cite{guo2021pi,Moon_2019_ICCV_3DMPPE, HMOR2020}, poses and shape~\cite{zanfir_cvpr_2018_multiple, zanfir_multipeople_sensing, jiang2020_multiperson, BMP_2021_CVPR, Fieraru_2020_CVPR}, or poses and motion~\cite{AgudoTPAMI2019,AgudoCVPR2017}. For instance, Jiang \textit{et al.}~\cite{jiang2020_multiperson}, have  considered global constraints accounting for inter-person occlusions, interpenetration between meshes and depth ordering. 

Nevertheless, while showing very promising results, previous approaches are still prone to fail in situations where people in the images have different heights. This is  because the constraints considered so far are not able to handle the inherent body size / depth ambiguity, and result in consistent body 2D reprojections, but wrong relative scales and depths (see Fig.~\ref{fig:teaser}-center). 

\begin{figure}
  \vspace{-0.45cm}
  \includegraphics[width=\linewidth, trim={0.1cm 0.1cm 0cm 0}, clip = true]{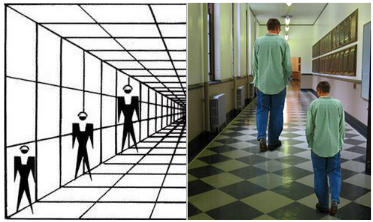}
  \vspace{-0.5cm}
\caption{Two examples of the Ponzo illusion~\cite{ponzom} on the perception of the human height. In both cases, the contextual information of the ground floor helps to perceive the height and depth of each person.}
\label{fig:ponzo}
\end{figure}

In this paper we present a novel approach for multi-person reconstruction that handles the depth/body size ambiguity. For this purpose, we will consider a novel constraint that estimates body scale and relative camera-body translation while enforcing the feet of all people to remain on the ground.
This observation, does indeed have psychological groundings on the geometrical-optical Ponzo illusion~\cite{ponzom}, that suggests that the  human mind judges an object's size based on its background, in our case the ground floor, and the preconceived prior that humans have their feet on the ground (see Fig.~\ref{fig:ponzo}).

In order to implement this constraint we present a simple but effective coarse-to-fine strategy where we first use~\cite{jiang2020_multiperson, totalCapture_frank} to extract an initial estimation of the shape and pose of all people in the image. This estimate may contain gross errors, like people floating in the air or with large discrepancies in their true size (see again Fig.~\ref{fig:teaser}-center).  We then extract the ground plane from the image using an off-the-shelf depth~\cite{midas_ranftl2021} and semantic segmentation estimation network~\cite{detectron2}, and compute the 3D normal direction of that plane. Finally, we devise a learning scheme, in which translation and scale parameters per person are optimized so as to minimize the feet-to-plane 3D distance and the 2D reprojection error. We observed that with only these two terms, more computationally demanding losses preventing interpenetrability of the 3D meshes were not necessary. 

We exhaustively evaluate the proposed approach on the MuPoTS-3D~\cite{singleshot} and 3DPW~\cite{3DPW} datasets, and show a consistent improvement over a depth ordering metric in comparison to recent state-of-the-art. To further stress the significance of our results, we propose two additional metrics that assess the accuracy of the retrieved body sizes and the inter-person distances. In both these metrics we also obtain remarkably better results than previous methods. Fig.~\ref{fig:teaser}-right, shows an example output of our approach where we are able to estimate correct body shape and position of two people with very different sizes.

\begin{figure*}[t!]
  \vspace{-0.45cm}
  \includegraphics[width=\linewidth, trim={0.1cm 0.1cm 0cm 0}, clip = true]{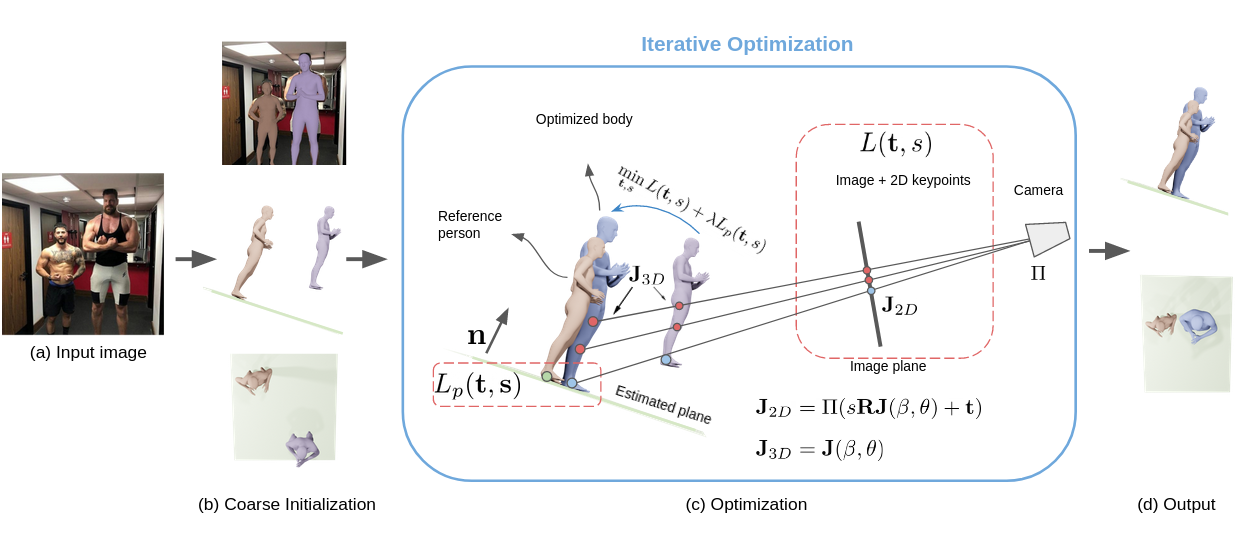}
  \vspace{-0.8cm}
  \caption{\small{\textbf{Overview of our approach.} Given an input image (a), we first estimate the 2D keypoints and SMPL parameters as the initialization (b). Then we estimate the ground plane normal and minimize the reprojection error along with the ground plane constraint for reconstruction (c). We are able to correct the spatial arrangement in the reconstructed 3D scene by forcing all body meshes to touch the ground with their feet (d). Notice the difference with respect to the results from the initial estimation where people are floating in the air (b).    
  }}
  \label{fig:model}
  \vspace{-0.4cm}
\end{figure*}


\section{Related work}\label{sec:sota}

\paragraph*{3D Human Pose Estimation.}
Estimating the 3D pose of human joints has received growing interest mostly for its importance in real-world applications and by the release of large-scale public MoCap and multi-camera datasets~\cite{ionescu_human3.6m:_2014, MPI-INF-3DHP, panoptic, pumarola_3dpeople:_2019}. A lot of important studies have been done on this topic for single-person~\cite{li20143d, tekin2016structured, Bogo:ECCV:2016, mehta2017vnect, Zhou_2017_ICCV,MorenoCVPR2017, pavlakos2017coarse,Simo_ijcv2017,yang20183d,  Zhao_2019_CVPR} and multi-person ~\cite{lcr, lcr++, 3dmppe, kumarapu2020animepose, singleshot}. 3D pose estimation for single-person is done directly from the image or lifting 2D joints to 3D joints and for multi-person either by using top-down or bottom-up approaches.
To face the shortcomings of loss of information due to reprojection in 3D pose estimation, parametric models of the human body have been extensively used. Examples of these are SMPL~\cite{loper_smpl:_2015}, Frank~\cite{totalCapture_frank} or  GHUM~\cite{GHUM_CVPR2020}. Although it is very difficult to annotate large-scale datasets with these models, there are optimization-based methods such as~\cite{Bogo:ECCV:2016, smplx_2019, fang2021mirrored} that allow to fit the parameters very well using only 2D and 3D joint annotations. 
Due to its simplicity and robustness several works~\cite{kanazawa_hmr, VIBE, pavlakos_learning_2018, neural_body_fitting_omran2018, smplx_2019, kolotouros2019cmr, Holopose_Guler_2019_CVPR, kolotouros2019spin, zanfir_cvpr_2018_multiple, hkmr, Zhang_2020_CVPR, zeng20203d} build upon the SMPL model to reconstruct human bodies from single images. The SMPL model can be used to recover the full body shape (as a 3D mesh) and joints.

\paragraph*{Multi-person 3D Pose and Shape Estimation.}
Recently there have been some works that extend 3D pose/shape estimation to a multi-person setting. The most relevant works in this area are ~\cite{zanfir_cvpr_2018_multiple, zanfir_multipeople_sensing, jiang2020_multiperson, BMP_2021_CVPR, fang2021mirrored, ROMP}. 

Zanfir~\textit{et al.}~\cite{zanfir_multipeople_sensing} are the first ones to use a bottom-up model to estimate the pose and shape of multiple-people. In a concurrent but radically different work, Zanfir~\textit{et. al}~\cite{zanfir_cvpr_2018_multiple} propose a top-down multi-staged approach. They first estimate initial pose for each person and then jointly optimize to absolute poses for multiple people in the image. They also include a ground plane constraint. However, they estimate it using initial 3D joint predictions and require a video sequence to do so. In contrast to them, we cannot assume that our initial 3D joints estimations are correct and we work with a single image. So we take a very different approach for estimating the ground plane normal. While these two works have been one of the firsts to address this multi-person problem, subsequent methods such as~\cite{jiang2020_multiperson} show improved results over them.  

Very recent approaches~\cite{BMP_2021_CVPR, ROMP} propose a novel mesh representation to account for cases of occlusions and also reason about the most probable depth location for each person in the scene while presenting bottom-up architectures. Jiang \textit{et al.}~\cite{jiang2020_multiperson} leverage the architecture proposed by HMR~\cite{kanazawa_hmr} to get multi-person shape/pose estimates. They include two novel losses: one that penalizes collisions between people and another depth ordering aware loss that uses segmentation masks to enforce a correct depth order. Due to the nature of this loss, the depth order can be misestimated when no people overlap with each other. As we will see, our method is capable of facing this limitation.

\paragraph*{Human-Scene Interaction.}
Recently, increasing attention has been given to the context of the scene when estimating the body pose and shape along with the spatial arrangement of humans in the scene. So far very few works~\cite{zhang2020phosa, Fieraru_2020_CVPR, prox_Hassan_2019_ICCV, Hassan:CVPR:2021} have been done in this area. For example, PHOSA~\cite{zhang2020phosa}, which is closer to our approach, seeks to model several people in a scene and their interactions with object on images on-the-wild. They propose a two stage approach where they first estimate each person and object in the scene and then globally optimize for spatial arrangement. Instead of objects, we use the ground plane which unifies all people in the scene.

\section{Method}\label{sec:method}

We next describe our approach (see Figure~\ref{fig:model} for an overview). 
First, we define the mesh representation adopted (Sec.~\ref{subsec:human_mesh}). Then our problem formulation and objective function are presented in Sec.~\ref{subsec:formulation}. Finally, we explain our feet-to-ground constraint (Sec.~\ref{subsec:feet_constraint}) and optimization scheme (Sec.~\ref{subsec:optimization}).

\subsection{SMPL and Pose/Shape Estimation}\label{subsec:human_mesh}

We adopt SMPL~\cite{smpl} to represent the 3D body geometry. This model is specially appealing in our context given its ability to represent a large number of shape and pose variations with a few number of parameters. In particular, SMPL  encodes the body geometry in two vectors: $\shape \in \mathbb{R}^{10}$, containing the parameters modelling the shape, and $\pose \in \mathbb{R}^{72}$, which encode the information regarding the person pose. Using previous definitions, SMPL also implements a function:
\begin{equation}
    M (\shape, \pose): \pose \times{\shape} \mapsto \mat{V} \in \mathbb{R}^{3B} \label{eq:smpl},
\end{equation}
which estimates a set of $B$ vertices $\mat{V}$ of the body mesh from the pose and shape parameters. Additionally, the model also provides a linear mapping $\mathcal{J}$ so that $\joints = \mathcal{J} (\mat{V})$ which allows to estimate the position of $K$ body joints $\joints \in \mathbb{R}^{3K}$ from the mesh vertices $\mat{V}$. In the following sections, we will refer to  the i-th joint of a body mesh with parameters $\pose$ and $\shape$ as $\joints (\shape, \pose)_{i}$.
 
Along with the body mesh, a weak-perspective camera $\Pi=[\sigma, t_{x}, t_{y}] \in \mathbb{R}^3$ is parametrized to project the mesh into image coordinates. To position the humans in the 3D space, the weak-perspective camera can be converted to the perspective camera projection by assuming a fixed focal length $f$ for all images, where the depth of the person is determined by the relation $d=f/\sigma$ and the absolute translation is given by:
\begin{equation}
    \trans = [t_{x}, t_{y}, d]
    \label{eq:smpl}.
\end{equation}

Given a set of 2D body joints $\jointsTwoD \in \mathbb{R}^{2K}$ in pixel coordinates and a confidence value $c_i$ associated to each of them, the pose and shape parameters defining the body geometry can be estimated by solving the following optimization problem:
\begin{align} 
&\min_{\shape,\pose,\rot,\mathbf{t},s} \sum_{i}{c_{i} || \jointsTwoD_{i} - \Pi(\rot \joints (\shape, \pose)_{i} + \mathbf{t})|| }+ L_{p} \label{eq:reproj_loss} 
\end{align}
where 
$\rot$ is a global rotation matrix and $L_{p}$ a regularizer.
Intuitively, the defined problem attempts to find the pose and shape parameters which  minimize the distance between the estimated 2D joints and the projection of the 3D mesh in  image coordinates.

\begin{figure}[t!]
  \vspace{-0.45cm}
  \includegraphics[width=\linewidth, trim={0.1cm 0.1cm 0cm 0}, clip = true]{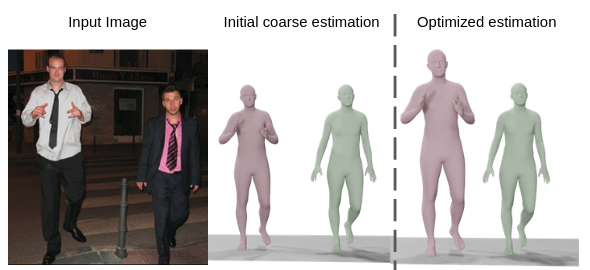}
  \vspace{-0.3cm}
  \caption{\small{\textbf{An illustration of the difference in body sizes}. The reconstructed 3D bodies are placed at the same depth and on the same plane. 
  Our method scales body sizes accordingly. In contrast, other state-of-the-art methods struggle capturing the variation of body sizes. 
  }}
  \label{fig:explanation}
  \vspace{-0.4cm}
\end{figure}

\subsection{Problem Formulation}
\label{subsec:formulation}
Consider an input RGB image $ {\bf I} \in \mathbb{R}^{H \times W \times 3} $ with $N$ persons
and $({\pose}^{1:N},{\shape}^{1:N})$ to be the set of SMPL parameters obtained by solving Eq.~\eqref{eq:smpl_optimization} for each individual. Our goal is to obtain the set of translation and scale parameters $({\trans}^{1:N},{\scale}^{1:N})$ for each person, which best explain the scene and produces feasible results with correct depths and sizes.

In order to obtain a set of translation and scale parameters leading to realistic results, we propose to solve a similar minimization problem to the one introduced in Eq.~\eqref{eq:smpl_optimization}. In our case, however, we introduce two important modifications. Firstly, we jointly optimize the individual parameters for each person by defining the loss as:
\begin{equation} 
    \lossRep({\trans}^{1:N},{\scale}^{1:N}) = \sum_{i,n}{c^n_{i} || \jointsTwoD^n_{i} - \Pi(s^n\rot^n \joints (\shape^n, \pose^n)_{i}
    + \mathbf{t}^n)||},
    \label{eq:multi_optimization}
\end{equation}
where $\rot^n$ is the estimated rotation matrix for the person $n$ and $(\jointsTwoD^n_{i},c^n_{i})$ is the $i$-th 2D joint in image coordinates and its associated confidence, respectively. Note that here, we introduce an additional scale parameter $s$ that modifies the overall size of the body (including height) without the need to change $\shape$.

Secondly, we introduce a regularizer conditioned on the scale and translation parameters that aims to minimize the feet-to-ground distance and avoid body configurations flying on the air. We will detail this loss in the following subsection.

Finally, we jointly find the optimal translation and scale for each person by minimizing:
\begin{align} 
    &\min_{\mathbf{t}^{1:N},s^{1:N}}  \lossRep({\trans}^{1:N},{\scale}^{1:N}) + \lambda \lplane({\trans}^{1:N},{\scale}^{1:N}),
    \label{eq:our_optimization}
\end{align}
where $\lambda$ is a hyper-parameter controlling the trade-off between the minimized projection error and the regularization term.

\subsection{Scale and Translation aware Feet-to-Ground Constraint} \label{subsec:feet_constraint}

The inclusion of the scale and translation aware regularizer in Eq.~\eqref{eq:our_optimization} is one of the key contributions of our paper. This is motivated by the observation that in most real scenes, the depicted individuals are typically standing on a common ground. While this observation has already been exploited in~\cite{zanfir_cvpr_2018_multiple}, the ground-to-plane distance in this work is conditioned uniquely on the 3D joints of the feet. In contrast, we condition this distance using also the body size scale and translation vector. 
By doing this we make it possible to address the depth-size ambiguity problem. To illustrate this, consider Figure~\ref{fig:model} (b). Given an estimated normal vector of a plane representing the scene ground, the estimation of the translation and scale parameters obtained with Eq.~\eqref{eq:smpl_optimization} produces unrealistic results where one of the persons is "flying".  However, this ambiguity can be addressed by   our optimization scheme, in which we  translate each of the individuals and change the size of their bodies so as to get close to the ground plane, while still minimizing the 2D reprojection error.

\vspace{1mm}
\noindent\textbf{Regularizer definition}: Assume that, for a given image, we have access to the vector $\mathbf{n}$ defining the normal of the ground plane. To enforce all people in the scene to touch the ground, we use:
\begin{align}
   \lplane&=\sum_{n}{|(\ankl^{n}(\mathbf{t}^{n}, {s}^{n}) - \mathbf{p})\cdot \normal| +  |(\ankr^{n}(\mathbf{t}^{n}, {s}^{n}) - \mathbf{p})\cdot \normal|}
      \label{eq:loss_plane}
\end{align}
where $\ankl^n$ and $\ankr^n$ are the left and right ankle 3D joints of the $n$-th person obtained after translating and scaling its mesh using the estimated parameters $\mathbf{t}^{n}$ and ${s}^{n}$. Being $k$ the index of these joints in $\joints (\shape^n, \pose^n)$, they are computed as:
\begin{equation}
\mathbf{x}_k=s^n \rot \joints (\shape^n, \pose^n)_{k} + \mathbf{t}^n  . 
\end{equation}
Additionally, $\mathbf{p}$ is a reference point fixed in the plane which is chosen as the ankle of a reference person, see Figure~\ref{fig:model}(c). Notice that we purposefully translate the plane to $\mathbf{p}$. We select the reference person as the one with lower initial reprojection error.

\vspace{1mm}
\noindent\textbf{Ground Normal Estimation:} Restricting and determining a person's movement in the 3D space can be possible with a correct ground plane normal $\mathbf{n}$. If, on top of this, we force the reprojected body joints to match a set of reference 2D keypoints, we can easily estimate correct translation and scale parameters. Unfortunately, we typically do not know the direction of $\mathbf{n}$ in real scenarios and thus, it must be estimated. For this purpose, we first obtain a depth map $ \bf{\widehat{{d}}} \in \mathbb{R}^{H \times W}$ by applying an off-the-shelf depth estimator to the input image~\cite{midas_ranftl2021}.
Then, we use panoptic segmentation~\cite{panoptic_seg} to identify the pixels corresponding to the ground regions in the image. Finally, we take the depth values from these ground regions and, using their corresponding $x,y$ pixel coordinates, obtain a set of 3D points. With these points we fit a plane, acquiring its normal $\normal$. To handle outliers we use the RANSAC~\cite{ransac}.

To enforce the condition for Eq.~\eqref{eq:loss_plane} that $\mathbf{p}$ must be a fixed point in the plane, we assume that $\mathbf{p}$ is a reference point with known depth. So we take $\mathbf{p}$ as a point in the 3D space and translate the entire plane to this point. All subsequent translation estimations are, therefore, relative to this point.

\subsection{Optimization}\label{subsec:optimization}
The loss and regularization terms minimized in Eq.~\eqref{eq:our_optimization} are fully-differentiable w.r.t the optimized variables. Therefore, to solve our optimization problem we use a gradient descent approach. In particular, we initialize $\trans^{1:N}$ with the estimations obtained using Eq.~\eqref{eq:smpl_optimization} and set all $\scale^{1:N}$ to one (mean scale). We iteratively minimize the objective by computing the gradients of the translation and scale parameters.

\section{Experiments}\label{sec:experiments}

\subsection{Datasets}\label{subsec:datasets}

In our experiments, we validate our method conducting exhaustive experiments over two standard benchmarks for human pose/shape estimation.

\vspace{1mm}
\noindent\textbf{MuPoTS-3D}~\cite{singleshot}: It is a multi-person dataset providing 3D ground truth for people in the scene. We use the same test sequences as in~\cite{jiang2020_multiperson}.

\vspace{1mm}
\noindent\textbf{3DPW}~\cite{3DPW}: It is a multi-person in-the-wild dataset, which features diverse motions and scenes. It contains 60 video sequences with 3D joints and translations annotations. We use this dataset to especially test our models capability of generality of applicability to challenging in-the-wild scenarios. We use only the test sequences that present the ground truth for more than one person as we focus on multi-person reconstruction.

\subsection{Metrics} \label{subsec:metrics}
In order to evaluate the quality of the multi-person 3D pose and shape estimations, we use three different metrics: (i) pairwise person's depth order ($d_{ord}$), (ii) pairwise normalized distances between persons in the scene ($d_{norm}$) and (iii) pairwise person's height discrete comparison ($h_{ord}$). The first metric was introduced in previous works~\cite{jiang2020_multiperson, BMP_2021_CVPR} and quantifies the percentage of correctly estimated ordinal depth relations between all pairs of people in the image. The other two metrics are proposed by us and allow to conduct a more thorough evaluation. 

\vspace{1mm}
\noindent \textbf{Pairwise normalized distances $d_{norm}$:} This metric complements the depth order  by measuring how accurate the distance relations are between pairs of people in the image. Thus, measuring the quality of the estimated translations. This metric is motivated by the observation that an estimation can have correct depth order for all people but have unrealistic or disproportional distances between them. If $d_{ord}$ is correct but each person is very far from each other, in comparison to what is perceived in the image, then the estimation cannot be considered accurate. The opposite also holds true, if $d_{norm}$ has a low score, it does not necessarily mean that the depth order is correct. Formally, this metric is defined as follows: 
\begin{align} 
    d &=  \sum_{i}\sum_{j}{\|(\mathbf{t}_{i} - \mathbf{t}_{j})\|}\label{eq:dist} \\
     d_{norm} &= \frac{\sum_{f=0}^{m} (d_{f} / d_{f}^{max} - \hat{d_{f}} / \hat{d_{f}^{max}}) } {m} \label{eq:distance_metric},
\end{align}
where $m$ is the number of frames and $i$ and $j$ are indices for the $i$-th and $j$-th persons in the frame $f$. Also, $d_{f}$ corresponds to the pairwise distances of the ground truth values and $\hat{d_{f}}$, to estimated ones, which are both calculated with Eq.~\ref{eq:dist}. Here, $\trans$ is the estimated translation of each person in the image. For each frame we normalize all distances by the maximum pairwise distance ($d_{f}^{max}$).

\vspace{1mm}
\noindent \textbf{Discrete height comparison $h_{ord}$:} To determine the quality of the estimated body scales, we use people's height as a proxy measure. One can reason that if a model is able to correctly estimate the height of each person, providing a good reprojection, then a correct spatial arrangement is more likely to occur. This metric is the percentage of correctly estimated ordinal height relations between all pairs of people in the image, i.e., smaller to bigger or vices-versa. This is very similar to $d_{ord}$ but applied to heights. This metric helps us determine that the scale has correctly increased or decreased a person's size in order to provide a better estimation. The heights are calculated by the euclidean distance across the limbs connecting the head and feet joints.

Overall, these three metrics attempt to measure the aspects of multi-person reconstruction that our method focuses on improving: translation and scale.

\begin{table*}
\centering
    \begin{tabular}{c|c|c|c|c|c|c}
  \cmidrule(lr){1-7}
  \multicolumn{1}{c}{\textit{}} & \multicolumn{3}{c}{\textit{MuPoTS-3D}}& \multicolumn{3}{c}{\textit{3DPW}}\\
  \cmidrule(lr){2-4}   \cmidrule(lr){5-7}
    
    \tb{Method} &   $d_{ord}$ $\uparrow$ & $d_{norm}$ $\downarrow$ & $h_{ord}$ $\uparrow$ & \
    $d_{ord}$ $\uparrow$ & $d_{norm}$ $\downarrow$ & $h_{ord}$ $\uparrow$\\    
    
    \midrule
       FrankMocap~\cite{rong2021frankmocap} &  85.56  & 0.492 & 52.79\  
       & 81.61   & \tb{0.463}      & 50.53       \\
      
        Ours (w/ FrankMocap) &  88.03 & 0.367 & 49.65 \   
        &  82.69    & {0.557}  & 51.60    \\

        CRMH~\cite{jiang2020_multiperson} & 92.20 & 0.351 & 52.43 \   
        & 76.13     & 0.737     & 51.32     \\
        
       Ours (w/ CRMH) & \textbf{95.58} & \textbf{0.243} & \textbf{57.75} \
       & \textbf{85.00}     & 0.516     & \textbf{55.80} \\

       BMP*~\cite{BMP_2021_CVPR}       & 94.50  & -    & -   &     - & -   & - \\

  \bottomrule
\end{tabular}
  \caption{\textbf{Results on representative datasets}. We present the results on three state-of-the-art methods and use two of them as initialization for our approach. Evaluation is done on MuPoTS-3D and 3DPW datasets with the metrics described in Sec.~\ref{subsec:metrics}. *The values were copied directly from the paper as there is no publicly code nor data available. }
    \label{tab:results}
\end{table*}

\subsection{Implementation details}
We jointly optimize for both translation and scale using the ADAM optimizer~\cite{adam} with learning rate 1e-2 for a total of 600 iterations. As reference 2D keypoints, we use the 3D reprojected joints obtained from the initial SMPL estimations from either CRMH~\cite{jiang2020_multiperson} or FrankMocap\cite{rong2021frankmocap}. We present experiments with these two initializations in Sec.~\ref{sec:experiments}, however, given the nature of our method, any other initialization can be used.  We use semantic segmentation from Detectron2~\cite{detectron2} for our baseline and use MiDasv3.~\cite{midas_ranftl2021} as depth estimator. For all cases we use a fixed focal length of $f=1000$.

\begin{table}
\centering
    \resizebox{0.45 \textwidth}{!}{%
    \begin{tabular}{c|c|c|c}
  \cmidrule(lr){1-4}
  \multicolumn{1}{c}{\textit{}} & \multicolumn{3}{c}{\textit{MuPoTS-3D}}\\
      \cmidrule(lr){2-4}
    
   \tb{Method} & $d_{ord}$ $\uparrow$ & $d_{norm}$ $\downarrow$ & $h_{ord}$ $\uparrow$  \\
       
    \midrule
       Only Reprojection    & 92.51                 & 0.415 & 55.48 
       \\
       Only Plane           & 95.10                 & 0.293 & 51.14
       \\
       Ours w/ GT-J2D       & 94.62                 & 0.318 & \textbf{59.28}
       \\
       Ours                 & \textbf{95.58}        & \textbf{0.243} & 57.75
        \\
    
  \bottomrule
\end{tabular}}
  \caption{\textbf{Ablative study for proposed losses}. We present the effect of each individual loss (reprojection, plane) and the effect of using ground truth 2D keypoints as inputs to the model instead of detected ones by and off-the-shelf detector.}
    \label{tab:ablation_res}
\end{table}

\subsection{Comparison with simple baseline}\label{subsec:baseline}
In this first experiment, we evaluate the effectiveness of the proposed ground-plane regularization. For this purpose, we compare our method with a baseline that incorporates depth information but does not explicitly model the scene ground. Concretely, our baseline estimates the depth of each person in the scene. This is done by averaging the depth values of all the pixels belonging to a person's mask (obtained with from semantic segmentation). Then, we optimize for $s$ and $\trans$ only over the $x$ and $y$ axis while maintaining $z$ equal to the estimated depth using Eq.~\ref{eq:multi_optimization} as objective function.

\begin{table}
\centering
  \resizebox{0.35 \textwidth}{!}{%
    \begin{tabular}{c|c|c|c}

  \cmidrule(lr){1-4}

    \multicolumn{1}{c}{\textit{}} &   \multicolumn{3}{c}{\textit{MuPoTS-3D}}\\
    \cmidrule(lr){2-4}

     \tb{Method} & $d_{ord}$ $\uparrow$ & $d_{norm}$ $\downarrow$ & $h_{ord}$ $\uparrow$     \\   
    
    \midrule
       Baseline   & 87.34       & 0.635      & 55.09      \\
       Ours       &  \tb{95.58} & \tb{0.243} & \tb{57.75}\\

  \cmidrule(lr){1-4}
  \multicolumn{1}{c}{\textit{}} &   \multicolumn{3}{c}{\textit{3DPW}}\\
    \cmidrule(lr){2-4}
    
     \tb{Method} & $d_{ord}$ $\uparrow$ & $d_{norm}$ $\downarrow$ & $h_{ord}$ $\uparrow$    \\   
         
    \midrule
       Baseline         & 82.24       & \tb{0.508} & 54.27     \\
       Ours             &  \tb{85.00} & 0.516      & \tb{55.80}\\
    
  \bottomrule
\end{tabular}}
  \caption{\textbf{Comparison with our baseline}. As a sanity check we compare our method with the baseline described in Sec.~\ref{subsec:baseline}.}
    \label{tab:baseline}
\end{table}

Comparing ourselves to a method that directly uses people's depth information is important as one part of our goal is to improve depth order. We can see in Table~\ref{tab:baseline} that our method outperforms this baseline by a significant margin. This justifies the choice of using the ground plain as a constraint. It is more reliable to estimate the ground plane's normal and use it as a constraint than to recover individual person's depth from an estimated depth map. The baseline relies on segmentation masks for computing a person's depth. In many cases these segmentation masks are not able to be recovered, especially under the presence of occlusions. This results in incorrect depth estimations. Thus, the poor performance of the baseline. In contrast, segmenting the ground is easier which facilitates the depth estimation of the ground plane and consequently the normal $\mathbf{n}$. For this reason, our method is able to make a  better use of depth information from a given depth map to improve 3D human pose-shape recovery.

\begin{figure*}[t!]
  \vspace{-0.1cm}
  \includegraphics[width=\linewidth, trim={0.1cm 0.1cm 0cm 0}, clip = true]{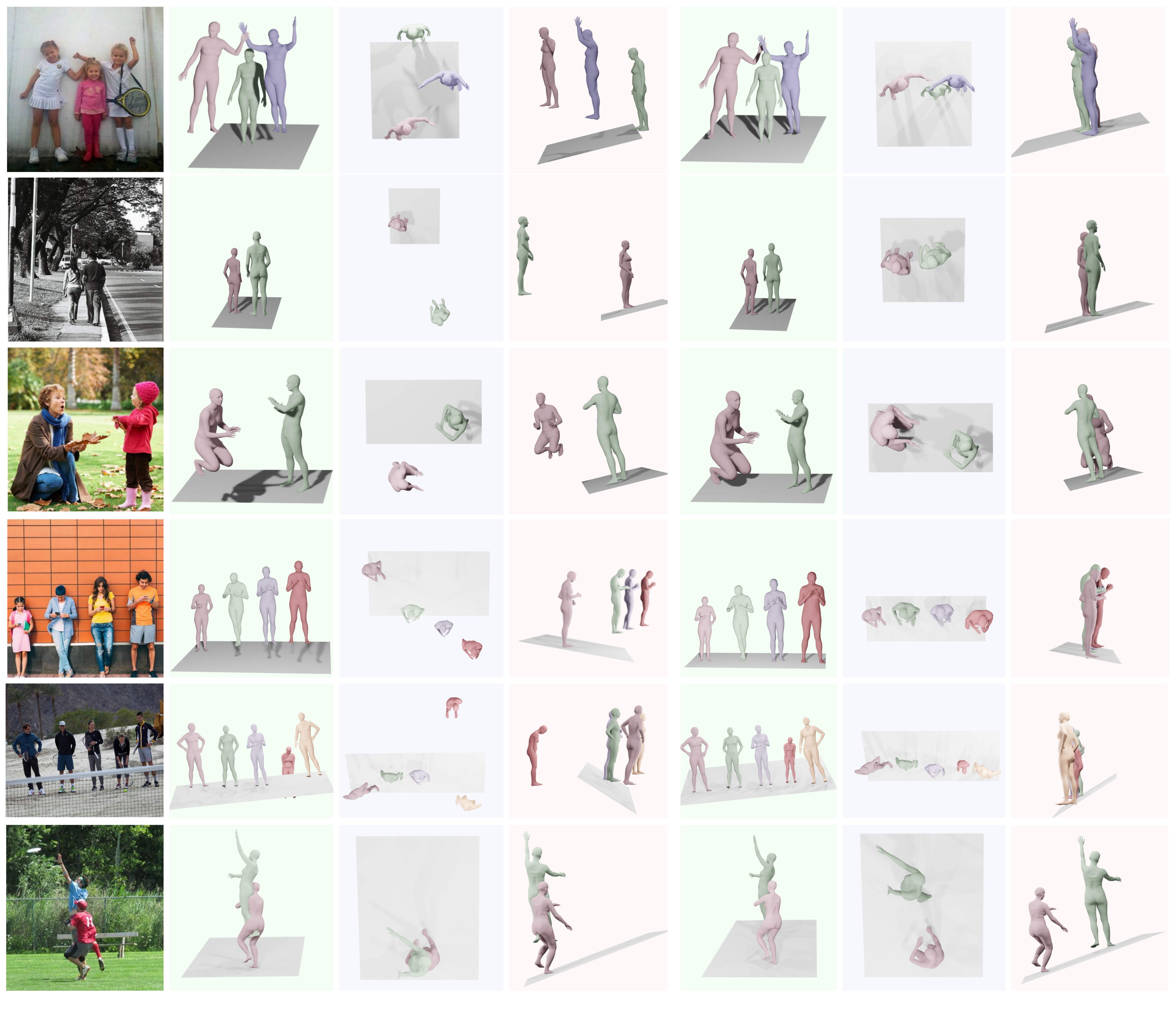}
        \put(-485,4){{{Input image}}}
        \put(-338,4){{{CRMH~\cite{jiang2020_multiperson}}}}
        \put(-115,4){{{Ours}}}
  \vspace{0.2cm}
  \caption{\small{\textbf{Qualitative results}. We visualize the results of our method from different viewpoints: front (green background), top (blue background), and side (red background) views. We compare the results with CRMH~\cite{jiang2020_multiperson} and improve depth order and overall scene coherence. Note that CRMH does not estimate a plane. We add the same plane estimated by our method for the visualization purposes.
  }}
  \label{fig:qualitative}
  \vspace{-0.4cm}
\end{figure*}

\subsection{Comparison with the state-of-the-art}\label{subsec:quantitative}

We compare our method to other multi-person approaches with regards to spatial arrangement, specially for correct depth order and body size and leave out any human pose comparison as we do not change the pose from the initial estimations. We measure the quality of the translation and scale estimations as they are crucial for facing the depth-size ambiguity.

We compare ourselves with two state-of-the-art methods: CRMH~\cite{jiang2020_multiperson} and BMP~\cite{BMP_2021_CVPR}, and an additional baseline for which we apply FrankMocap~\cite{rong2021frankmocap} to the multi-person setting. The results in both datasets (MuPoTS-3D, 3DPW) are presented in Table~\ref{tab:results}.

\vspace{1mm}
\noindent\textbf{Improving depth order.} We first evaluate our proposed method for depth order and correct translation of body meshes in the 3D space, as they are both related. 
The results are shown in the first two columns of each dataset in Table~\ref{tab:results}. It can be seen that our approach is able to improve the spatial arrangement of the initial methods used (\cite{rong2021frankmocap} and \cite{jiang2020_multiperson}). Concretely, it can be seen that our method improves CRMH by 3.38\% (95.58\% vs. 92.20\%) and FrankMocap by 2.47\% (88.03\% vs. 85.56\%) in the $d_{ord}$ metric for the MuPoTS-3D dataset and CRMH by 8.87\% (85.00\% vs. 76.13\%) and FrankMocap by 1.08\% (82.69\% vs. 81.61\%) in 3DPW dataset, which contains images in-the-wild. We also improve $d_{norm}$ score by a good margin. For example, in MuPoTS-3D we improve CRMH's score by 0.108 (0.243 vs. 0.351) and FrankMocap by 0.125 (0.367 vs. 0.492). A similar improvement is presented in 3DPW dataset when using CRMH as initial estimation. Our method also outperforms the current state-of-the-art~\cite{BMP_2021_CVPR} by 1.08\% in the depth order metric. Please note that BMP~\cite{BMP_2021_CVPR} results are taken from the original paper and not recomputed by us as we do for the other methods as their code and data is not yet publicly available.

\vspace{2mm}
\noindent\textbf{Relative body sizes.} After stabilising that we can estimate more accurate translation parameters, we now focus on the people's body sizes, which also play an important role when trying to correctly perceive the 3D information of the scene. In the same fashion as before, we measure the correct height order ($h_{ord}$), meaning the scale relations (bigger/smaller) between people in a scene. Here we find that our method improves CRMH by 5.32\% (57.75\% vs. 52.43\%) in MuPoTS-3D, CRMH by 4.48\% (55.80\% vs. 51.32\%) in 3DPW and FrankMocap by 1.07\% (51.60\% vs. 50.53\%)in 3DPW. A lower score on this metric shows that there are estimations where people that should be bigger in comparison to others in the scene, are actually estimated as smaller and vice-versa.

\subsection{Ablation studies}\label{subsec:ablation}

In Table~\ref{tab:ablation_res} we analyse the effect of using only one of the terms of our objective function. First, we use only the reprojection term for the optimization and, later, we use only the feet-ground constraint. We do this to validate that both of the terms contribute to the overall performance of the model. Additionally, we use "perfect" 2D keypoints as input, taken from the ground truth, to see if there is an upper-bound limit to our method's performace and how this affects the results. Note that a discussion on the effects and limitations of our feet-ground constraint is presented in the Supplementary Material.

It can be seen that using only reprojection helps significantly to estimate better height results as $h_{ord}$ is relatively high in comparison to using only the ground plane constraint. However, this term has no contribution with the depth order nor correct proportional distance between people, having the worst scores on $d_{ord}$ and $d_{norm}$, just as expected. It can be seen that using only the feet-ground constraint alone leads to better depth ordering and better $d_{norm}$ score than when using only reprojection. This is expected as we first move the people towards the plane in the camera direction during optimization, meaning that reprojection is partially conserved even without the reprojection term. As we can see in the results, heights estimations here are the worst as $h_{ord}$ scores lower, which is also expected.

We also perform an ablation study that uses ground truth 2D keypoints annotations used by the reprojection loss. As it turns out, our method surprisingly yields to better results in the $d_{ord}$ and $d_{norm}$ metrics than by using ground truth 2D keypoint annotations. However, this setting yields to the best results in height order ($h_{ord}$) showing that better reference keypoints results in better height estimates but not necessarily into better depth ordering.

\subsection{Robustness to initialization}\label{subsec:depth}
To show that our framework can work along with any other initial method, we present results in both datasets using FrankMocap~\cite{rong2021frankmocap} estimations as initial point. Note that this method is originally trained for single-person estimation. The results presented in Table~\ref{tab:results} show a good improvement over~\cite{rong2021frankmocap} used directly as a multi-person approach. Our method is also robust to 2D keypoint initialization as it has been established in the previous section.

\subsection{Qualitative Results}\label{subsec:qualitative}
We present our qualitative results in Figure~\ref{fig:qualitative}. 
As expected, our method produces results with more realistic spatial arrangement while capturing the diversity of body sizes. This yields improved depth ordering and better proportional distances among people in the scene when compared to method~\cite{jiang2020_multiperson}. In all cases shown, we use CMRH for initial estimations.  The image in the last row of Figure~\ref{fig:qualitative} presents a failure case. Here, one person is jumping and the ground constraint is not completely met. Although, in this case, the depth order is still improved by our method, the distance between each person may not be accurate. More qualitative results, including failure cases, and discussion of the effect of the ground plane constraint can be found in the Supplementary Material.

\section{Conclusions}\label{sec:conclusion}

In this paper, we have presented a novel optimization scheme in order to address the inherent depth/size ambiguity in multi-person 3D body pose estimation. In particular, our method jointly computes the optimal translation and scale parameters of all the individuals in the scene by imposing a constraint forcing them to stand into a shared ground-plane. In order to estimate this information, we use off-the-shelf depth-map and semantic segmentation methods to automatically extract the ground-plane normal from the image. Comparing our approach to state-of-the-art methods over benchmark datasets, we show that the proposed pipeline provides more coherent solutions in terms of the relative scale and translation of the estimated persons. More concretely, we are able to consistently improve the performance by a considerable margin in multiple metrics evaluating the accuracy of the pose parameters. Finally, it is worth mentioning that our method is model agnostic and only requires a set of initial estimations for the 3D joints. Therefore, it can be combined with any previous or future method for human mesh reconstruction from still images.

\noindent{\bf Acknowledgements}: 
\small{
This work is partly supported  by the Spanish government with the project MoHuCo PID2020-120049RB-I00,  the ERA-Net Chistera project IPALM PCI2019-103386 and María de Maeztu Seal of Excellence MDM-2016-0656. Adria Ruiz acknowledges financial support from MICINN (Spain) through the program Juan de la Cierva.
}

{\small
\bibliographystyle{ieee}
\bibliography{references}
}

\end{document}